\documentclass[11pt]{article}

\usepackage[final]{acl}

\usepackage{times}
\usepackage{latexsym}

\usepackage[T1]{fontenc}

\usepackage[utf8]{inputenc}

\usepackage{microtype}

\usepackage{inconsolata}

\usepackage{graphicx}

\usepackage{makecell}
\usepackage{tabularx} 
\usepackage{pifont} 
\usepackage{xcolor} 
\usepackage[table, dvipsnames]{xcolor} 
\newcommand{\greencheck}{{\color{green}\ding{51}}}
\usepackage{pifont} 
\usepackage{xcolor} 
\newcommand{\xmark}
{{\color{red}\times}}
\usepackage{multirow}
\usepackage{multicol}
\usepackage{amsmath}
\usepackage{tcolorbox}
\usepackage{listings}

\newtcolorbox{promptbox}[1][]{colback=gray!5,colframe=gray!75,fonttitle=\bfseries,title=Prompt,#1}
\title{EgoErrorVQA: Assess Egocentric Comprehension Capabilities through Procedural Errors for Ego-Agentic AI}

\author{Junlong Li\textsuperscript{\rm 1}, Junxi Li\textsuperscript{\rm 1}, Jianjun Gao\textsuperscript{\rm 2}, Chen Cai\textsuperscript{\rm 2}, Lap-Pui Chau\textsuperscript{\rm 1}, Yi Wang\textsuperscript{*,\rm 1}\\
\\
\textsuperscript{1} Department of EEE, The Hong Kong Polytechnic University \\
\textsuperscript{2} School of EEE, Nanyang Technological University\\
  \textsuperscript{*}\small Corresponding author\\
  \texttt{\small junlong.li@connect.polyu.hk},      \texttt{\small yi-eie.wang@polyu.edu.hk} \\
  }

\begin{document}
\maketitle
\begin{abstract}The majority of our everyday activities are procedural and consist of sequences of interdependent steps. However, existing benchmarks for Visual Agents and Visual Language Models (VLMs) overlook the evaluation of their procedural comprehension ability from an egocentric visual perspective, particularly for detecting procedural errors, a critical capability for everyday assistance. To bridge this gap, the EgoErrorVQA task is firstly proposed for egocentric procedural comprehension with explicit procedural errors modeling. Besides, we develop a user-friendly evaluator agent based on the Agent2Agent (A2A) protocol, enabling rigorous and standardized evaluation of visual agents through VQA-based interaction. A range of models are evaluated using both open-ended and multiple-choice questions, revealing persistent weaknesses in handling procedural errors and error types. Moreover, we introduce Ego-ADR, an Adaptive Decoupled Reasoning framework that decouples complex procedural reasoning to enhance models’ understanding of procedural errors. It achieves consistent performance gains over the selected baselines and attains state-of-the-art results on several metrics under comparable settings. Code: \url{https://github.com/z1oong/EgoErrorVQA}
\end{abstract}

\section{Introduction}

Egocentric video understanding is particularly important as it captures the world from a first-person perspective \citep{plizzari2024outlook} and underpins agentic and embodied applications \citep{fung2025embodied}. A key yet under-evaluated requirement in this setting is procedural understanding \citep{li2025building,li2026egoprocevqa}: many everyday tasks involve executing interdependent steps under ordering constraints (e.g., assembling a toy car), which demands step-level comprehension, and temporal reasoning beyond recognition or captioning \citep{ging2024open}. Moreover, robust assistance requires not only following procedures but also detecting failures \citep{flaborea2024prego}, such as out-of-order steps, omissions, wrong-object usage, and redundant actions, which is crucial for enabling agents to provide effective support in smart homes and smart factories.

During a procedural task like making a sandwich (take bread → spread mayo → add ham → close sandwich), procedural error detection requires an agent to recognize step omissions (forgetting the ham), out-of-order execution (closing before adding ham), or wrong-object usage (grabbing the wrong condiment). However, existing benchmarks fall short in several respects. First, despite the growing interest in egocentric procedural understanding \citep{bansal2022my}, procedural errors are rarely characterized in a way that enables systematic evaluation. Also, existing procedural error related datasets typically contain only task-specific error labels, which are difficult to transfer to other scenarios, and the area lacks a systematic definition of procedural error types. Second, conventional benchmarks are not well aligned with the emerging visual agents and video-LLMs, error detection based solely on existing datasets lacks interaction between visual and textual information, and evaluation setup does not faithfully reflect VQA scenarios in which AI assistants usually perform to support people in everyday life.

\begin{table*}
    \centering
    \vspace{-23pt}
    \scalebox{0.7}{
    \begin{tabular}{cccccccccc}
      \hline
         Benchmark & Videos&QA-pairs&   Procedural&   \makecell{Aciton\\Label}&  \makecell{Error\\Centric}&  \makecell{LLM\\Scoring}&  \makecell{Multiple\\Scenarios}& \makecell{Open-\\end\\Question}&\makecell{Multiple-\\choice\\Question}\\
         \hline
         EgoVQA \citep{fan2019egovqa} & 16& 600&  $\xmark$&   $\xmark$&   $\xmark$&  $\xmark$& \greencheck &  \greencheck&\greencheck\\
         AssistQ \citep{wong2022assistq} & 100& 531&  \greencheck&\greencheck &   $\xmark$&  $\xmark$&  \greencheck& $\xmark$ &\greencheck\\
         EgoTaskQA \citep{jia2022egotaskqa} & 2K&40K&  \greencheck& \greencheck &   $\xmark$&  $\xmark$&  \greencheck&  $\xmark$&\greencheck\\
         EgoPlan-Bench \citep{chen2023egoplan} &-&4939&  \greencheck&  \greencheck&   $\xmark$&  $\xmark$&  \greencheck&  $\xmark$&\greencheck\\
         EgoSchema \citep{mangalam2023egoschema} & -&5063&  $\xmark$&  $\xmark$&   $\xmark$&  $\xmark$&  \greencheck&  $\xmark$&\greencheck\\
         VidEgoThink \citep{cheng2024videgothink} & 195&600&  $\xmark$&  \greencheck&   $\xmark$&  \greencheck&  \greencheck&  \greencheck&$\xmark$\\
         OpenEQA \citep{majumdar2024openeqa} & -&1636&  $\xmark$& $\xmark$ &   $\xmark$&  \greencheck&  \greencheck&  \greencheck&$\xmark$\\
         ProMQA \citep{hasegawa2025promqa} & 384&401&  \greencheck&  \greencheck&  $\xmark$ &  \greencheck&  $\xmark$&  \greencheck& $\xmark$\\
          EgoTextVQA \citep{zhou2025egotextvqa} & 1507&7064&  $\xmark$& $\xmark$ &   $\xmark$&  \greencheck&  \greencheck&  \greencheck&$\xmark$\\
         \hline
         EgoErrorVQA (Our) & 800&5417&  \greencheck&  \greencheck&    \greencheck&  \greencheck&  \greencheck&  \greencheck&\greencheck\\
         \hline
    \end{tabular}
    }
    \caption{Comparison between EgoErrorVQA and common video benchmarks. Videos represents the number of original videos.}
    \vspace{-10pt}
    \label{tab1}
\end{table*}

To address these limitations, we propose EgoErrorVQA, a novel task that evaluates visual agents' and video-LLMs' egocentric video comprehension, and reasoning capabilities by formulating procedural error detection and classification as a VQA task. It contains an eight-category error taxonomy covering nearly all common procedural error types. Besides, we create procedure texts that serve as an instructional description for each type of procedural task, and propose an evaluator agent (green agent) that interacts with the agent under test (white agent) via a unified Agent2Agent (A2A) protocol\footnote{More details about A2A protocl are available at \url{https://a2a-protocol.org/latest/}} : the evaluator serves information such as the video path together with both open-ended questions (e.g., action correctness) and multiple-choice questions (e.g., error-type identification), collects the model’s responses, and automatically produces evaluation results. In practice, users only need to establish an A2A connection between the target model and the evaluator agent on a public platform, upon which the benchmark assessment is automatically executed end-to-end, greatly improving usability and reproducibility. Moreover, we propose an adaptive decoupled reasoning framework that achieve substantial performance improvements in procedural understanding. In summary, our main contributions are as follows:

• Unlike generic procedural understanding focused on a single scenario, we introduce EgoErrorVQA, a novel task which provides the first agentic procedural error detection and classification benchmark across diverse scenarios as shown in Table~\ref{tab1}. Experiments reveal that current models still lag behind humans in those tasks.

• We propose EgoErrorVQA-E as an evaluator agent with fixed interfaces to streamline reproducible evaluation of external agents and video-LLMs, and integrate a scoring suite that combines human-aligned LLM-based judgment (open-end) with general evaluation metrics (multiple-choice). It assesses not only semantic quality, but metric-based performance, making the evaluation more comprehensive and broadly applicable.

\begin{table}
    \centering
    \vspace{-6pt}
    \scalebox{0.6}{
    \begin{tabular}{ccccccc}
    \hline
         Benchmark& Source &Clips& \makecell{Correct\\samples}&\makecell{Error\\samples} &\makecell{Q-Len\\$/$A-Len}&  \makecell{QA\\pairs}\\
         \hline
         \multirow{4}{*}{Open-end}& Cap& 960&  632& 328 &\multirow{4}{*}{12.86$/$15.57}&  \multirow{4}{*}{3560}\\
         &  Oops&215&  175& 40 &&  \\
         &  Tent&184&  124& 60 &&  \\
         &  Assem&446&  341& 105 &&  \\
         \hline
         \multirow{4}{*}{Multiple-choice}& Cap& 1000&  660& 340 &\multirow{4}{*}{-$/$1.38}&  \multirow{4}{*}{1857}\\
         &  Oops& 215& 175& 40 &&  \\
         & Tent&182&  123& 59 &&  \\
         & Assem&460&  352& 108 &&  \\
         \hline
    \end{tabular}
}

    \caption{Statistics of EgoErrorVQA. Cap, Oops, Tent, Assem represents CaptainCook4D, EgoOops, Epic-tent, and Assembly101. Q-Len, A-Len is the average length of questions and answers, measured in words.}
    \vspace{-13pt}
    \label{tab2}
\end{table}

• As a simple yet effective attempt, we propose Ego-ADR, decoupling procedural understanding processes to enhance the model’s capacity for procedural comprehension. Under comparable conditions, it achieves state-of-the-art performance on key metrics, offering insights for developing vision-based agent skills in procedural tasks.

\begin{figure*}
    \centering
    \vspace{-20pt}
    \includegraphics[width=1\linewidth]{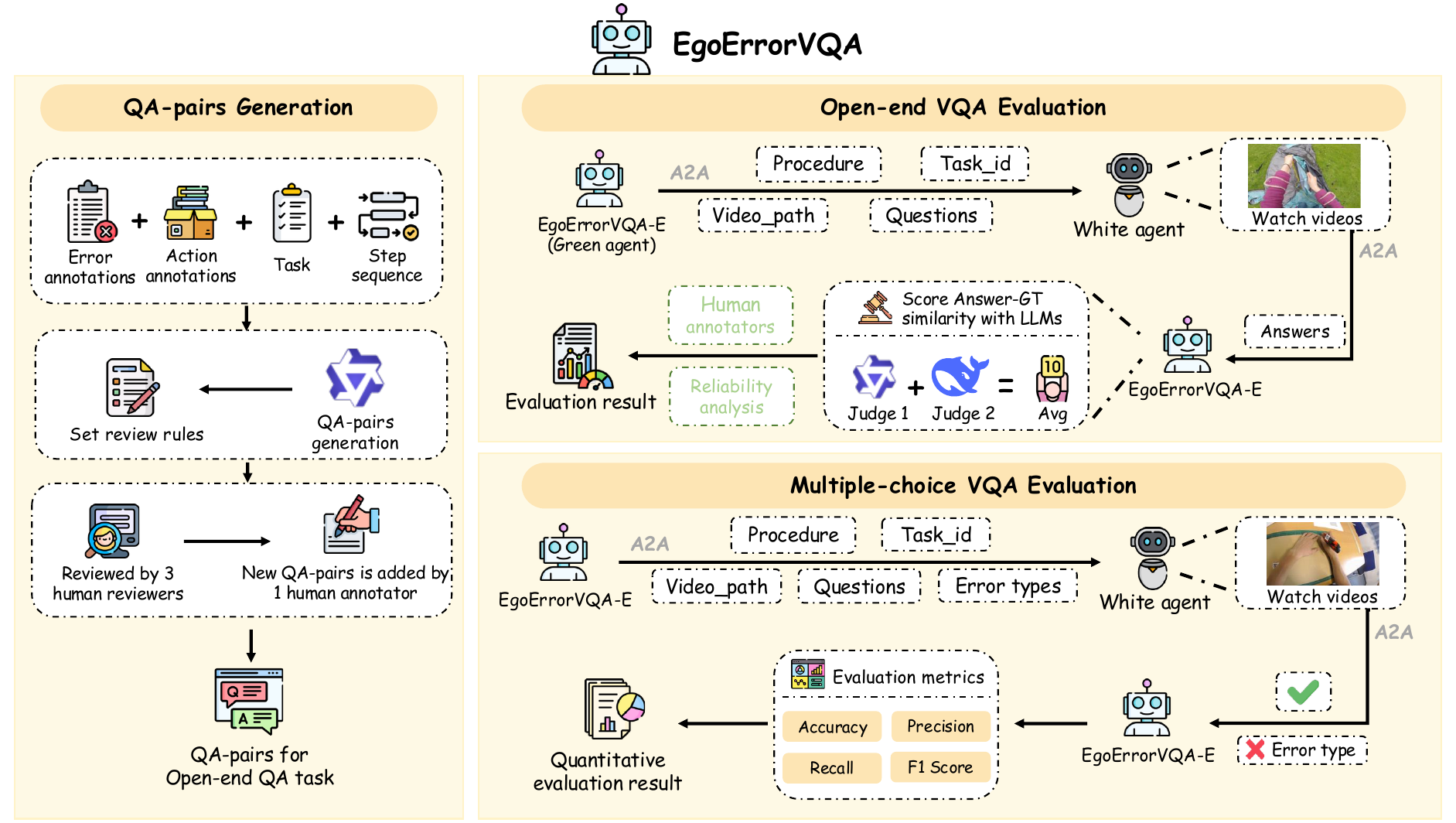}
  \vspace{-18pt}
    \caption{Overview workflow of EgoErrorVQA. In the QA-pairs Generation stage, it outlines the overall procedure for constructing QA-pairs and highlights the stages and roles of human involvement. Subsequently, the Open-end VQA Evaluation and Multiple-choice VQA Evaluation sections illustrate the overall pipelines for those two evaluation tasks, respectively.}
    \vspace{-10pt}
    \label{fig1}
\end{figure*}

\section{Related Work}

\textbf{Egocentric video benchmark.} Existing egocentric video benchmarks for VQA models typically involve simple questions or tasks such as action recognition and object identification \citep{cheng2024egothink,majumdar2024openeqa,jia2022egotaskqa}, offering only short-term assessments of reasoning and recognition. Some works consider more complex tasks like task planning \citep{chen2023egoplan}, which better probe reasoning and memory. Although some works \citep{hasegawa2025promqa} have begun to construct evaluations on procedural tasks, such as cooking activities, our work is uniquely focused on egocentric procedural error detection and classification, a task that is more challenging and insightful, covering 31 types of procedural activities in 4 types of scenarios.

\textbf{Egocentric procedural error detection.} As egocentric datasets proliferate \citep{haneji2025egooops,peddi2024captaincook4d}, downstream tasks have become increasingly diverse, with egocentric procedural error detection \citep{wang2023holoassist} emerging as a key challenge. Such task is central to enabling AI assistants \citep{li2025building} to effectively support daily activities and even help blind people. Our work differs from existing error-annotated datasets in that we systematize procedural errors across a wide range of everyday scenarios using a proposed error taxonomy, thereby establishing a unified and comprehensive evaluation framework. In addition, by constructing a VQA dataset, we evaluate video-LLMs and agents’ understanding of procedural errors from multiple perspectives, in a way that simulates how users pose questions and how AI assistants respond, further supports the research on interpretable error detection and classification.

\begin{figure*}
    \centering
    \vspace{-20pt}
    \includegraphics[width=1\linewidth]{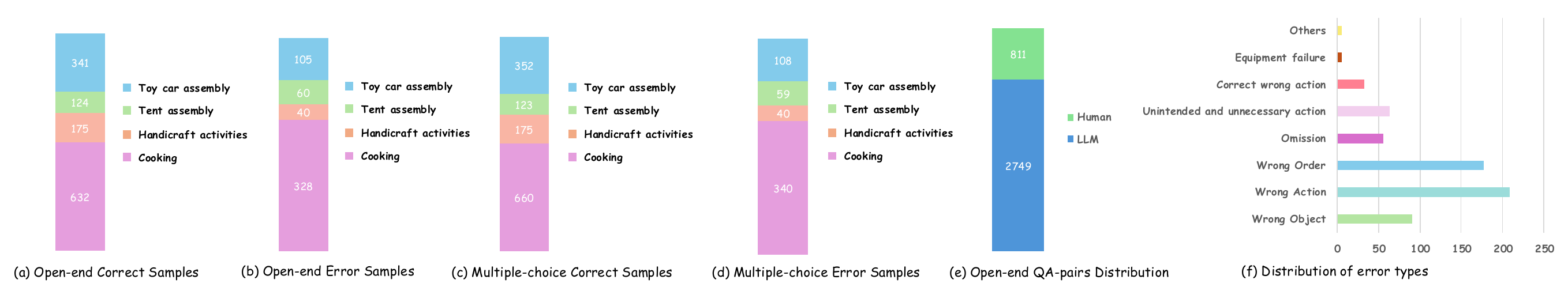}
\vspace{-15pt}
    \caption{(a), (b), (c), and (d) respectively show the proportions of correct and incorrect samples in open-end VQA and multiple-choice VQA. (e) shows the proportion of QA-pairs in the open-end VQA that are generated by the LLM versus those added by human annotator. (f) shows the distribution of each error type.}
    \label{fig2}
\end{figure*}

\section{Open-end VQA in EgoErrorVQA}

Given that video question answering (VQA) is already a core capability of current agents and video-LLMs, and is highly likely to remain so in the future, we design our evaluation procedure in the form of VQA. To enable a more comprehensive and objective evaluation, EgoErrorVQA incorporates both open-end and multiple-choice VQA as shown in Fig.~\ref{fig1}. In this section, we focus on the open-ended setting.

\subsection{Data Collection}

In Table~\ref{tab2}, we select four egocentric procedural task datasets: \textbf{CaptainCook4D} \citep{peddi2024captaincook4d}, \textbf{EgoOops} \citep{haneji2025egooops}, \textbf{Epic-Tent} \citep{jang2019epic} and \textbf{Assembly101} \citep{sener2022assembly101}. CaptainCook4D covers 24 cooking recipes; EgoOops includes five handicraft tasks (e.g., electrical circuits, ionic reactions); Epic-Tent focuses on tent setup; and Assembly101 on toy car assembly. Considering potential future expansion of the training set, we subsample each dataset proportionally to its size and categorize the data into four scenario types, yielding 1,000 samples for \textbf{Cooking} (CaptainCook4D), 215 for \textbf{Handicraft activities} (EgoOops), 184 for \textbf{Tent assembly} (Epic-Tent), and 460 for \textbf{Toy car assembly} (Assembly101).

\subsection{Task Formulation}

In the open-ended setting, EgoErrorVQA-E first transmits to the evaluated White agent, via a communication protocol, a Procedure that outlines the complete workflow for the task, specifying the main steps required to achieve the goal. White agent is then asked to answer carefully designed questions that probe, from multiple perspectives, whether specific steps and their ordering are appropriate. The answers are sent back to EgoErrorVQA-E, which performs scoring with a brief rationale.

\subsection{QA-Pairs Generation}

Rigorous evaluation of a model’s understanding of procedural tasks cannot rely on generic questions such as “Is there anything wrong in this video?” Because the video clips contain a lot of redundant information, the agent struggles to focus on critical procedural elements. These responses thus provide little evidence of hierarchical procedural understanding or stepwise logical reasoning, and do not support meaningful evaluation. Targeted and deliberately confounding questions about specific actions and steps are therefore required to obtain data that more accurately reflects the agent’s understanding of procedures and procedural errors.

For example, for a correctly executed step, a question such as “Did I add chopped cilantro to the ramen bowl as instructed?” directly targets that action and requires the model to reason about it. Similarly, one might ask “Did I use the wrong part when attaching the roof to the body?”, while the actual error occurs elsewhere in the video, thereby introducing a controlled distractor.

We adopt an \textbf{LLM–Human collaboration strategy} for QA-pair generation. \textbf{Qwen2.5-7B-Instruct} \citep{bai2025qwen2} first labels each procedural step by task type and existing annotations, then generates 2-3 diverse QA-pairs per sample. Around 4000 QA-pairs are independently checked by three human annotators over 80 hours, who remove or revise those that fail to test procedural error understanding or focus on irrelevant content. A final annotator adds QA-pairs for missing evaluation aspects and introduces deliberately misleading questions about correctly executed steps (e.g., “Did I add any unnecessary steps when I attach engine to chassis?”). Details can be found in Appendix~\ref{a1}.

In total, EgoErrorVQA contains 3,560 QA-pairs covering 1,805 samples, with each sample associated with 1–3 QA-pairs. Approximately 2,749 QA-pairs are generated by an LLM and subsequently refined through human review, while about 811 are manually authored, shown in Fig.~\ref{fig2}.

\begin{figure*}
\vspace{-20pt}
    \centering
    \includegraphics[width=1\linewidth]{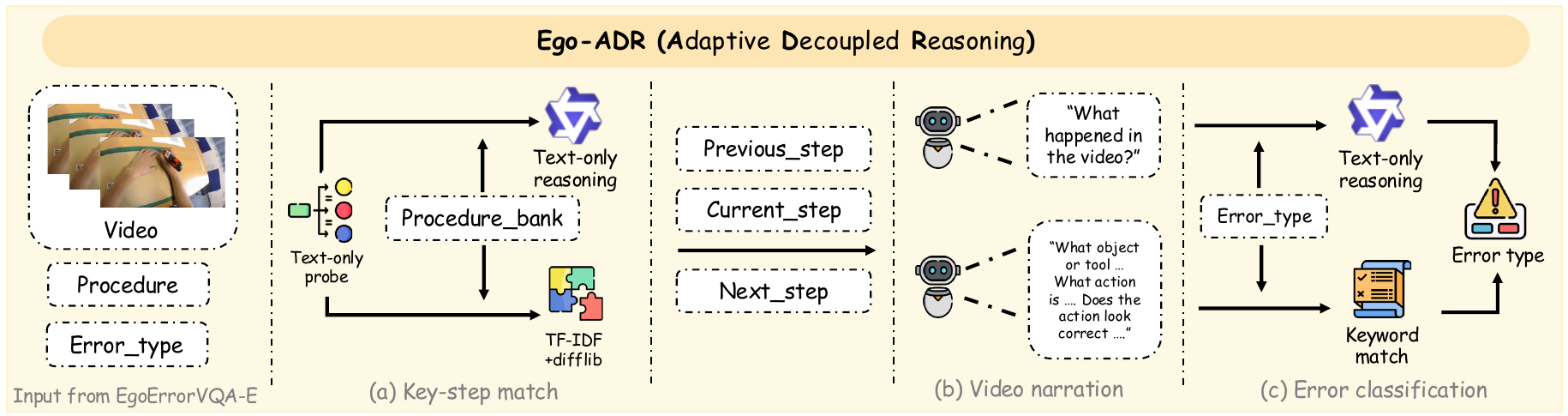}
    \caption{Overall Architecture of Ego-ADR. In steps (a), (b), and (c), the upper path represents deep decoupling, whereas the lower path represents shallow decoupling.}
    \label{fig3}
\end{figure*}

\subsection{Evaluation Metric}

Given the extensive evidence supporting the effectiveness of LLM-as-a-Judge \citep{li2025generation}, we adopt a novel scoring scheme in our evaluation. To facilitate reproducibility and broader adoption, we employ two open-source, memory-efficient LLMs, \textbf{Qwen2.5-7B-Instruct} \citep{bai2025qwen2} and \textbf{DeepSeek-LLM-7B-Chat} \citep{bi2024deepseek} as judges.

To ensure reliability and mitigate subjective bias, the judging models do not perform complex reasoning over the answers. Instead, they only assess the semantic similarity between the model-generated answer and the ground truth. Also, we design a rigorous reliability analysis of LLM scoring in subsequent sections. To further reduce model-specific bias, both Qwen and DeepSeek provide Sim., representing semantic similarity, and we take their average as the final score. Sim. is rated on a 0–5 scale, where \textbf{5 = Perfect match}, \textbf{4 = Minor error}, \textbf{3 = Partially correct}, \textbf{2 = Mostly wrong}, \textbf{1 = Wrong}, and \textbf{0 = No response}.

\section{Multiple-choice VQA in EgoErrorVQA}

To enable a fair and quantitative comparison between current agents and traditional egocentric procedural error detection methods, we introduce a multiple-choice VQA task that more rigorously evaluates models’ understanding and recognition of error categories, thereby mitigating the key limitation of interpretability in downstream error detection and classification.

\subsection{VQA Setting}

We categorize the error types into the following eight classes: \textbf{Wrong Object}, \textbf{Wrong Action}, \textbf{Wrong Order}, \textbf{Omission}, \textbf{Unintended and Unnecessary Action}, \textbf{Correct Wrong Action}, \textbf{Equipment Failure} and \textbf{Others}. Details of error type classification can be found in Appendix~\ref{a4}.

Multiple-choice VQA supplies white agent with the task-specific procedures and error types' definitions. To limit distraction from irrelevant content, each video sample is annotated with action labels, and the agent is instructed only to determine whether an error occurs in the specified step and, if so, to identify its type. The evaluation set comprises approximately 1,857 samples and share the same data source with open-end VQA, covering 31 procedural tasks. The specific input and output settings are provided in Appendix~\ref{a3}.

\subsection{Evaluation Metric}

For a fair comparison with traditional egocentric procedural error detection approaches \citep{lee2024error} and to comprehensively assess the model’s understanding of different error categories, we compute \textbf{Precision}, \textbf{Accuracy}, \textbf{Recall}, and \textbf{F1 Score}, and report the corresponding confusion matrix. Also, \textbf{Accuracy} for each error type is calculated to analyze the source of the model’s deficiency.

\section{Ego-ADR}

Procedural tasks span diverse real-world scenarios, and training on scenario-specific data cannot adequately improve model performance and may even hinder transfer to other contexts. Therefore, in Fig.~\ref{fig3}, we propose Ego-ADR, an \textbf{A}daptive \textbf{D}ecoupled \textbf{R}easoning framework that enhances models’ procedural understanding capabilities in a zero-shot setting, to inform the design of video-based procedural agent skills.

\subsection{Overall Architecture}

Through egocentric procedural error detection and classification task with video-LLMs and agents, we find that the model bears a heavy inference burden: it must localize the current procedural step, check for violations in the video, and further categorize error types. To mitigate this single-pass load, we adaptively decouple the reasoning pipeline and let the model focus on its strengths in textual reasoning and visual description. Concretely, the task is factorized into three stages: \textbf{Key-step match} to localize the current step in procedural context given the queried action, \textbf{Video narration} to describe the video segment, and \textbf{Error classification} to detect and categorize errors by comparing the narration with the procedural context and error taxonomy. For models such as Video-LLaVA that cannot accept text-only input, deep decoupling is suboptimal; instead, we adopt a shallow decoupling strategy that better exploits their inherent single-pass reasoning capability. We design a text-only probe, as shown in Fig.~\ref{fig3}, to assess whether a model can accept text-only input and perform reasoning. For models that support it, Ego-ADR performs deep decoupling, details can be found in Appendix~\ref{a7}.

\subsection{Decoupling Levels}

\textbf{Deep decoupling.} We first apply Qwen2.5-VL-7B-Instruct to segment the entire procedure into a structured procedure bank, where each step is annotated with detected objects. In \textbf{Key-step match}, we use text-only reasoning to retrieve the current, preceding, and subsequent steps and prune redundant information from the procedural text. The model then converts visual content into a detailed textual description (\textbf{Video narration}). Finally, based on the narration, the matched procedural context, and error type definitions, the model performs text-only reasoning to determine the error category (\textbf{Error classification}). 

\textbf{Shallow decoupling.} For models that cannot perform text-only reasoning, we simplify the matching mechanism. In \textbf{Key-step match}, we construct a TF-IDF (Term Frequency-Inverse Document Frequency) index over all step descriptions in procedure bank using unigram and bigram features, and match the step with the highest cosine similarity. If the similarity falls below 0.25, we fall back to the Ratcliff / Obershelp pattern algorithm to match character-level sequence. In \textbf{Video narration}, models must be guided to avoid misclassification caused by divergent phrasings of the same semantics in next stage. Specifically, in the prompt we instruct the model to describe (1) the object or tool, (2) the action, and (3) the anomalous event. Finally, we construct error-type-specific vocabularies to assist the model in keyword matching to classify procedural errors (\textbf{Error classification}).

\begin{table}
  \centering
  \vspace{-19pt}
  \scalebox{0.58}{
  \begin{tabular}{ccccccc}
    \hline
   Method& frames& Cook&  Tent&  Toy&  Hand&  Avg-Sim.\\
   \hline
   \textbf{Human}&-& 3.79& 3.89& 3.91& 3.47&  3.77\\
\hline
 \textbf{GPT-4o}&8f& 2.97& 3.14& 3.46& 3.5&  3.17\\
 \hline
   \textbf{GPT-4o-mini}&8f& 2.84& 3.18& 3.33& 3.33&  3.05\\
\hline
   \textbf{Gemini-2.5-flash}&8f& 2.82& 2.48& 2.89& 3.39& 2.87 \\
    \hline
   \rowcolor{blue!11}\multicolumn{7}{c}{\textbf{Open-Source 7B / 8B}}\\
    \hline
   \multirow{4}{*}{\textbf{LLaVA-OneVision}}&8f&  2.85&  3.17&  3.82&  3.34&  3.18\\
          &16f&  2.89&  3.19&  3.75&  3.35&  3.19\\
         &24f&  2.91&  3.25&  3.73&  3.37&  3.20\\
        &32f&  2.94&  3.22&  3.83&  3.42&  3.25\\
\hline
\multirow{3}{*}{\textbf{Vinci}} &8f& 2.98& \textbf{3.42}& 3.74& 3.61& 3.29\\
 &16f& 2.97& 3.36& 3.82& \textbf{3.64}& 3.30\\
 &24f& 2.97& 3.20&  3.70& \underline{3.63}& 3.25\\
 \hline
 \multirow{3}{*}{\textbf{EgoGPT}} &8f& 2.55& 2.96& 3.54& 3.24& 2.92\\
 &16f& 2.59& 2.89& 3.52& 3.21& 2.92\\
  &24f& 2.54& 2.92& 3.56& 3.21& 2.91\\
 \hline
\textbf{Video-LLaVA} &8f& 2.85& 3.12& 3.08& 3.04& 2.96\\
 \hline
 \multirow{4}{*}{\textbf{Video-LLaMA2}}&8f& 2.89& 2.41& 3.27& 3.16& 2.97\\
 &16f& 2.85& 2.67& 3.22& 3.16& 2.96\\
&24f& 2.86& 2.83& 3.18& 3.18& 2.97\\
&32f& 2.82& 2.59& 3.10& 3.14& 2.90\\
\hline
\multirow{4}{*}{\textbf{Qwen2-VL-7B-Instruct}}&8f& 3.01& \underline{3.37}& 3.83& 3.62& \underline{3.32}\\
 &16f& 3.02& 3.27& 3.81& 3.59& 3.31\\
&24f& 3.01& 3.26& \textbf{3.94}& 3.56& \textbf{3.33}\\
&32f& 3.01& 3.14& \underline{3.87}& 3.57& 3.30\\
\hline
\multirow{4}{*}{\textbf{Qwen2.5-VL-7B-Instruct}}&8f& 3.11& 2.90& 2.55& 3.53& 3.00\\
 &16f& \underline{3.14}& 2.95& 2.57& 3.47& 3.02\\
&24f& \textbf{3.18}& 3.05& 2.59& 3.50& 3.06\\
&32f& \underline{3.14}& 3.01& 2.62& 3.44& 3.03\\
\hline
\multirow{4}{*}{\textbf{Qwen3-VL-8B-Instruct}}&8f& 3.03& 2.75& 2.74& 3.48& 2.98\\
 &16f& 3.02& 2.85& 2.80& 3.56& 3.01\\
&24f& 3.00& 2.84& 2.80& 3.47& 2.99\\
&32f& 3.01& 2.85& 2.78& 3.56& 3.00\\
\hline
\rowcolor{blue!11}\multicolumn{7}{c}{\textbf{Open-Source 32B / 38B}}\\
    \hline
    \multirow{2}{*}{\textbf{Qwen3-VL-32B-Instruct}}&8f& 3.09& 2.70& 2.75& 3.28&2.99 \\
 &16f& 3.08& 2.94& 2.80& 3.29& 3.02\\
 \hline
 \multirow{2}{*}{\textbf{InternVL3.5-38B-Instruct}}&8f& 2.38& 2.54& 2.93& 2.95&2.60 \\
 &16f& 2.35& 2.49& 2.90& 2.87& 2.56\\
 \hline
  \end{tabular}
  }
  \caption{Models' performance on open-end VQA. \textbf{Cook} denotes results on Cooking, \textbf{Tent} on Tent assembly, \textbf{Toy} on Toy car assembly, \textbf{Hand} on Handicraft activities. \textbf{Avg-Sim.} denotes the sample-size-weighted average over all tasks. Details of Human performance shown in Appendix~\ref{a3}.}
  \vspace{-15pt}
  \label{tab3}
\end{table}

\begin{table*}
  \centering
  \vspace{-25pt}
  \scalebox{0.65}{
  \begin{tabular}{cccccccccccc}
    \hline
   Method& frames&  Acc&  Pre&  Recall& F1 &C&Om&UA&WA&WOb&WOr\\
   \hline
    \textbf{Human} &-&  85.9&  82.6&  65.5& 73.1&-&-&-&-&-&-\\
    \hline
     \textbf{GPT-4o} &8f&  57.9&  \textbf{17.5}& 25.6 & \textbf{20.8} &75.0&\underline{26.8}&\textbf{70.5}&9.6&23.3&13.6\\
    \hline
     \textbf{GPT-4o-mini} &8f&  28.5&  9.1&  45.5& 15.1&31.8&3.6&1.6&30.6&15.6&46.9\\
     \hline
      \textbf{Gemini-2.5-flash} &8f& 51.4& 11.1&  16.5& 13.3&68.3&8.9&32.8&13.4&13.3&12.4\\
    \hline
   \rowcolor{blue!11}\multicolumn{12}{c}{\textbf{Open-Source 7B / 8B}}\\
    \hline
   \multirow{4}{*}{\textbf{LLaVA-OneVision}} &8f&  64.5&  5.5&  3.3& 4.2&92.2&1.8&19.7&0&0&0\\
          &16f&  65.5&  8.4&  4.3& 5.7&92.2&1.8&29.5&0&0&0\\
         &24f& \underline{66.2} & 14.2& 7.1& 9.4&92.2&1.8&\underline{50.8}&0&0&0\\
        &32f&  65.2&  11.3&  6.3& 8.1&91.0&1.8&\underline{50.8}&0&0&0\\
\hline
\multirow{3}{*}{\textbf{Vinci}} &8f& 60.0& 6.4& 5.1& 5.7&84.1&0&1.6&8.6&3.3&6.2\\
 &16f& 59.5& 6.9& 5.9&6.4&83.1&0&1.6&8.6&3.3&6.2\\
 &24f& 58.8& 6.5& 5.6&6.0&82.2&0&0&10.1&3.3&7.3\\
 \hline
 \multirow{2}{*}{\textbf{EgoGPT}} &8f& \textbf{66.4}& 7.5& 2.5&3.8&\textbf{94.0}&0&19.7&0&0&0\\
 &16f& 65.9& 7.6& 3.0&4.3&\underline{93.1}&0&23.0&0&0&0\\
 \hline
 \textbf{Video-LLaVA} &8f& 44.2& 7.2& 16.3&10.0&58.7&0&0&22.0&18.9&12.4\\
 \hline
 \multirow{3}{*}{\textbf{Video-LLaMA2}}&8f& 18.4& 9.1& 62.4&15.8&15.4&0&4.9&48.3&17.8&45.8\\
 &16f& 21.9& 10.7& 53.6&17.8&19.3&0&0&56.9&20.0&46.9\\
&24f& 20.1& 10.8& 53.9&18.0&16.3&0&0&60.3&20.0&46.3\\
\hline
\multirow{4}{*}{\textbf{Qwen2-VL-7B-Instruct}}&8f&  17.5& 8.6& 73.9&15.5&14.3&0&0&51.7&\textbf{53.3}&22.6\\
 &16f&  18.4& 8.1& 70.3&14.6&16.5&0&0&46.9&\underline{52.2}&21.5\\
&24f& 20.5& 9.0& 72.0& 16.0&18.6&0&0&52.6&\underline{52.2}&21.5\\
&32f& 21.0& 8.9& 68.0&15.8&19.5&0&0&53.6&46.7&23.2\\
    \hline
\multirow{4}{*}{\textbf{Qwen2.5-VL-7B-Instruct}}&8f&  12.8& 10.1& \underline{97.3}&18.2&4.5&0&0&62.2&35.6&\underline{49.2}\\
 &16f&  13.3& 10.0& 96.7&18.1&5.4&0&0&63.2&37.8&45.2\\
&24f&  14.1& 10.6& \textbf{97.4}&19.0&5.8&0&0&64.1&40.0&48.6\\
&32f& 14.7& 11.0& 96.5&19.7&6.1&0&0&\underline{64.6}&38.9&\textbf{53.7}\\
\hline
\rowcolor{blue!11}\multicolumn{12}{c}{\textbf{Open-Source 32B / 38B}}\\
    \hline
    \multirow{2}{*}{\textbf{Qwen3-VL-32B-Instruct}}&8f&  39.5& 14.0& 39.8&20.7&45.3&19.6&0&62.3&26.3&30.0\\
 &16f&  41.7& \underline{14.5}& 35.9&\underline{20.7}&49.0&\textbf{28.6}&0&\textbf{67.2}&24.9&23.2\\
 \hline
 \multirow{2}{*}{\textbf{InternVL3.5-38B-Instruct}}&8f& 49.0& 8.1& 12.0&9.7&66.1&0&0&7.2&14.4&22.6\\
 &16f& 49.1& 7.8& 11.7&9.4&66.5&0&0&6.7&11.1&21.5\\
 \hline
 \rowcolor{gray!11}\textbf{Ours (Qwen2-VL)}&8f& 21.5 \textcolor{green}{$\uparrow${\scriptsize 22.9\%}}& 11.2 \textcolor{green}{$\uparrow${\scriptsize 30.2\%}}& 67.5&19.3 \textcolor{green}{$\uparrow${\scriptsize 24.5\%}}&17.6 \textcolor{green}{$\uparrow${\scriptsize 23.1\%}}&0&0&\cellcolor{gray!25}76.6 \textcolor{green}{$\uparrow${\scriptsize 48.2\%}}&31.1&27.1 \textcolor{green}{$\uparrow${\scriptsize 19.9\%}}\\
 \rowcolor{gray!11}\textbf{Ours (Qwen2.5-VL)}&8f& 19.6 \textcolor{green}{$\uparrow${\scriptsize 53.1\%}}& \cellcolor{gray!25}11.3 \textcolor{green}{$\uparrow${\scriptsize 11.9\%}}& 80.3&\cellcolor{gray!25}19.8 \textcolor{green}{$\uparrow${\scriptsize 8.8\%}}&13.9 \textcolor{green}{$\uparrow${\scriptsize 209\%}}&0&0& 67.0 \textcolor{green}{$\uparrow${\scriptsize 7.7\%}}& 45.6 \textcolor{green}{$\uparrow${\scriptsize 28.1\%}}& 45.2 \\
  \rowcolor{gray!11}\textbf{Ours (Video-LLaVA)}&8f & 13.6& 7.8 \textcolor{green}{$\uparrow${\scriptsize 8.3\%}}& 84.6 \textcolor{green}{$\uparrow${\scriptsize 419\%}}&14.3 \textcolor{green}{$\uparrow${\scriptsize 43\%}}&9.3&0&\cellcolor{gray!25}23.0 \textcolor{green}{$\uparrow$}&34.9 \textcolor{green}{$\uparrow${\scriptsize 58.6\%}}& \cellcolor{gray!25}55.6 \textcolor{green}{$\uparrow${\scriptsize 194\%}}&22.0 \textcolor{green}{$\uparrow${\scriptsize 77.4\%}}\\
  \hline
  \end{tabular}

}  
  \caption{Performance of multiple-choice VQA. C, Om, UA, WA, WOb, WOr means Accuracy of Correct, Omission, Unintended and Unnecessary Action, Wrong Action, Wrong Object, and Wrong Order, respectively. Other not mentioned error types are all zero. Green arrows indicate improvements and relative percentages over the baseline model. Gray bars show Ego-ADR results, where darker gray denotes state-of-the-art performance under the same parameter count and input frames. Details of Human performance shown in Appendix~\ref{a3}.}
  \vspace{-7pt}
  \label{tab4}
\end{table*}
\begin{table}
    \centering
    \vspace{-5pt}
  \scalebox{0.65}{ 
  \begin{tabular}{cccccc}
  \hline
         Method& Level & TP & Precision&  Recall& F1\\
         \hline
        \multirow{3}{*}{Qwen2-VL}& Deep & 172 \textcolor{green}{$\uparrow$}&  11.2 \textcolor{green}{$\uparrow$} & 67.5 \textcolor{green}{$\uparrow$} & 19.3 \textcolor{green}{$\uparrow$}\\
         & Shallow & 84 & 10.6 & 21.1 & 14.1 \\
         & CoT & 75 & 8.2 & 21.7 & 11.9\\
         \hline
        \multirow{3}{*}{Qwen2.5-VL} & Deep & 183 \textcolor{green}{$\uparrow$} & 11.3 \textcolor{green}{$\uparrow$} & 80.3 \textcolor{green}{$\uparrow$} & 19.8 \textcolor{green}{$\uparrow$}\\
         & Shallow & 76 & 7.1 & 22.7 & 10.8 \\
          & CoT & 134 & 8.2 & 77.9 & 14.8 \\
         \hline
 \multirow{3}{*}{Video-LLaVA} & Deep & 84 & 4.6 & 97.7 & 8.8\\
         & Shallow & 132 \textcolor{green}{$\uparrow$}& 7.8 \textcolor{green}{$\uparrow$}& 84.6 & 14.3 \textcolor{green}{$\uparrow$} \\
          & CoT & 86 & 5.3 & 61.0 & 9.8 \\
         \hline
    \end{tabular}
}    \caption{Ablation experiments to demonstrate the necessity of adaptive decoupling, where \textbf{Deep} denotes deep decoupling and \textbf{Shallow} denotes shallow decoupling. Green arrows indicate the best performance.}
    \label{tab5}
    \vspace{-8pt}
\end{table}

\section{Experiment}

\subsection{Experiment Setting}

\textbf{Baseline.}To conduct a comprehensive evaluation, we select three closed-source models: GPT-4o, GPT-4o-mini \citep{hurst2024gpt} and Gemini-2.5-flash \citep{comanici2025gemini}, three popular open-source video VLMs: LLaVA-OneVision \citep{li2024llava}, Video-LLaMA2 \citep{cheng2024videollama}, and Video-LLaVA \citep{lin2024video}, as representative models for general video understanding. Additionally, we include two visual agents fine-tuned on egocentric data, EgoGPT \citep{yang2025egolife} and Vinci \citep{10.1145/3749513}, to represent specialized egocentric vision agents. To support future research, we test three widely-used versions of Qwen-VL \citep{bai2025qwen2,wang2024qwen2,yang2025qwen3}, highlighting potential performance variations across iterations. Two larger models are evaluated, Qwen3-VL-32B-Instruct and InternVL3.5-38B-Instruct \citep{chen2024internvl}, demonstrating performance differences across parameter sizes.

\textbf{Benchmark Setting.} To assess the impact of frame sampling, we uniformly sample 8, 16, 24, or 32 frames within the action interval (if allowed by GPU memory), using 24 GB memory in 7B/8B evaluation.

\textbf{Ego-ADR Setting.} Based on text-only reasoning capabilities, we perform deep decoupling on \textbf{Qwen2-VL-7B-Instruct} and \textbf{Qwen2.5-VL-7B-Instruct}, shallow decoupling on \textbf{Video-LLaVA}, uniformly sample 8 frames as input, and keep all other settings identical to those in the previous experiments.

\subsection{Benchmark Results}

\textbf{Open-end VQA.} Table~\ref{tab3} shows that the highest Avg-Sim. 3.33 is obtained by Qwen2-VL-7B-Instruct (24-frame input), followed by Vinci (16-frame input) getting 3.30, whereas humans reach 3.77. This indicates only moderate performance on open-end VQA targeting specific actions, neither of the two closed-source models showed a clear advantage and increasing the number of input frames alone yields no substantial gains. However, given a maximum of five, the gap between 3.77 and 3.33 remains large, revealing significant limitations in procedural understanding. Analysis of raw outputs from strong reasoning models such as Qwen3-VL shows that their step-by-step reasoning is easily distracted by irrelevant details and selected large-parameter model InternVL3.5 produces abnormally terse responses, which harms performance on this task. Consequently, open-end results alone are insufficient to assess procedural error detection and must be interpreted jointly with multiple-choice outcomes.

\textbf{Multiple-choice VQA.} As shown in Table~\ref{tab4}, all models (even three closed-source models) perform poorly on procedural error detection and classification compared with humans, indicating substantial room for improvement. The highest accuracy, 66.4\% (EgoGPT, 8-frame input), reflects only moderate overall correctness. The highest precision, 17.5\% (GPT-4o, 8-frame input), indicates that current models struggle to accurately classify procedural errors and identify error types. The highest recall, 97.4\% (Qwen2.5-VL-7b-Instruct, 24-frame input), shows that most errors are detected, but mainly at the level of recognizing deviations from correct procedures rather than providing fine-grained error categorization. The best F1 score, 20.8\% (GPT-4o), further confirms that all models perform inadequately on this task. Meanwhile, human performance reached 85.9\%, 82.6\%, 65.5\%, and 73.1\% on these metrics, respectively. Models attain relatively high accuracy on more salient error types, such as Wrong Action, Wrong Object, and Wrong Order, but generally struggle with Omission and Unintended Action, likely due to insufficient global procedural comprehension. In contrast, GPT-4o, with its stronger reasoning capabilities, performs well on these two categories, reaching 70.5\% accuracy on Unintended Action. No model successfully detects Correct Wrong Action, Others, or Equipment Failure, likely reflecting both the intrinsic difficulty of these categories and their limited sample size. It is worth noting that no single metric adequately reflects model performance: relatively high accuracy may arise from class imbalance (e.g., predicting all samples as correct), while high recall alone only signifies error detection, and low precision reveals limited explanatory capability for the detected errors. Even if the model’s recall exceeds that of humans, markedly lower performance on other metrics merely indicates a tendency to over-label samples as errors under the given prompts, rather than a genuine understanding of procedural errors and error types.

Under identical inputs, agents show systematic biases: higher accuracy often coincides with lower recall, reflecting a tendency to label most samples as either correct or erroneous. Only GPT-4o and Qwen3-VL-32B-Instruct attain more balanced performance and thus higher F1 scores. Persistently low precision and weak Avg-Sim. further indicate that agents lack strong explanatory capacity for their predictions.

\begin{table}
    \centering
    \vspace{-16pt}
    \scalebox{0.7}{
    \begin{tabular}{cccccc}
    \hline
         Match&  Narration&   Precision&  Recall& F1\\
         \hline
          $\xmark$ & $\xmark$ &  10.1 & 97.3 & 18.2\\
         $\xmark$ & \greencheck &  11.1 & 78.7 & 19.4\\
        \greencheck & $\xmark$ & 12.4 & 24.3 & 16.5 \\
        \greencheck & \greencheck   & 11.3 & 80.3 & 19.8\\
        \hline
    \end{tabular}
}    
        \caption{Ablation experiments to demonstrate the effectiveness of decoupled reasoning, where \textbf{Match} denotes key-step match and \textbf{Narration} denotes video narration.}
    \label{tab6}
    \vspace{-16pt}
\end{table}

\subsection{Ego-ADR Results}

Through Ego-ADR framework, all three models achieved performance gains across multiple metrics, indicating that under adaptive decoupling the framework is effective not only for strong text-only reasoning models but also for Video-LLaVA, which does not accept text-only input. On the core metric Precision, Qwen2-VL, Qwen2.5-VL, and Video-LLaVA achieved relative improvements of 23.2\%, 10.6\%, and 7.7\%, respectively, with Qwen2.5-VL surpassing Gemini-2.5-flash to reach state-of-the-art performance among all 7B/8B models under the 8-frame input setting. Although Recall decreased except Video-LLaVA, the decline for the Qwen models is attributable to the removal of spuriously high Recall caused by misclassifying many samples as errors. This is corroborated by the increase in TP (True Positive) count, reflecting a more accurate understanding of error types. For the F1 score, which more comprehensively reflects error-classification capability, the three models improved by 19.7\%, 8.1\%, and 30.1\%, with Qwen2.5-VL again achieving state-of-the-art performance (in 7B/8B models). Furthermore, error-type recognition accuracy improved substantially, setting new state-of-the-art results for Wrong Action and Wrong Object. Notably, while Video-LLaVA was previously unable to identify Unintended Action errors, under Ego-ADR it achieved 23\% accuracy in this category.

\subsection{Ablation Studies}

\textbf{Necessity of Adaptive Decoupling.} Deep decoupling is appropriate for models with text-only reasoning abilities, as these models handle textual information effectively. In this setting, text reasoning improves the accuracy of key-step matching and error classification. In contrast, for Qwen under shallow decoupling, keyword-matching accuracy is highly sensitive to variation in descriptive style, causing performance degradation across four representative metrics. For Video-LLaVA, which lacks support for text-only reasoning, we approximated text-only reasoning by providing a completely black image alongside textual input. This led to declines in all metrics except Recall in Table~\ref{tab5}. Further analysis showed that the increased Recall arose from the model misclassifying many samples as errors. Although Recall improved, this behavior resembled guessing rather than reasoning, resulting in lower Precision and F1 scores. These findings indicate that forcing models without robust text-only reasoning capabilities to perform such tasks degrades their reasoning performance and underscores the need for adaptive decoupling.

\textbf{Effectiveness of Decoupled Reasoning.} This part is based on the deep decoupling architecture of Qwen2.5-VL. The key-step matching component reduces the model’s reasoning burden but yields only a 2\% relative improvement in overall metric F1 in Table~\ref{tab6}. By contrast, the narration stage is crucial: removing it causes the model to label most samples as correct, preserving high precision but reducing recall by 69.7\% and F1 by 16.7\%. Besides, Ego-ADR outperforms the widely used CoT (Chain-of-Thought), on all three models. Thus, describing before judging substantially enhances the model’s reasoning ability.

\subsection{Agreements between Human and Evaluators}

To assess the reliability of our LLM-as-a-Judge strategy, three human annotators and two judge models conducted the same scoring task. The Cohen’s Kappa among the annotators was 0.711, indicating strong inter-rater agreement. Human scores were then aggregated via majority voting to obtain a consensus label. The Pearson correlation between this consensus score and the judge scores was 0.851 for Qwen and 0.781 for DeepSeek, suggesting close alignment with human judgment and supporting the reliability of the judge models. Further experimental details are provided in Appendix~\ref{a6}.

\section{Conclusion}

To thoroughly evaluate egocentric procedural understanding of agents and VLMs, we propose EgoErrorVQA, a novel task featuring a new VQA benchmark dataset. Focusing on procedural error detection and classification that require in-depth reasoning, we design both open-end and multiple-choice evaluations to comprehensively assess agents’ understanding of procedural errors. We further propose an evaluator agent, enabling automated, dialogue-based assessment to simulate a realistic QA assistance scenario. Results suggest that agents still struggle with this task. Therefore, we propose Ego-ADR, decoupling complex procedural reasoning to enhance procedural errors understanding. We hope our work offers valuable insights for the relevant research fields.

\section*{Limitations}

Our dataset covers a wide range of scenarios and tasks but has several limitations. First, the four source datasets differ in size, so their contributions to the benchmark are not exactly equal. Meanwhile, the present work only involves evaluation data and does not include training data. Anticipating future training set expansion, we ensured that each dataset retains sufficient samples for extension. Because the original datasets contain many more correct than incorrect samples, the benchmark maintains a similar ratio, with correct samples dominating. This reduces score differentiation among models in open-end VQA. In multiple-choice VQA, accuracy and related metrics should therefore not be interpreted in isolation and require cautious analysis.

Besides, during detailed data annotation and experiments, we observed that the video dataset contains instances requiring fine-grained action and object recognition, such as choosing an incorrect heating time by pressing the small button on the microwave. Although these cases occur in only a small fraction of clips, they can still affect the model’s ability to classify error types. In Ego-ADR, we only conduct evaluations on multiple-choice VQA because this setting uses general quantitative metrics to clearly demonstrate the performance gains brought by our method, and avoids misjudgment caused by the inherent creativity of answers in open-end VQA.

\section*{Ethical Considerations}

We build our VQA dataset on publicly available egocentric datasets: CaptainCook4D, EgoOops, Epic-Tent, and Assembly101, all licensed for research use and all comply with ethical standards. We further verify that the constructed dataset contains no violent, illicit, or otherwise harmful content and does not disclose any private information. Besides, the annotators (one phd, two master students) involved in this work are all listed among the authors and have been compensated appropriately. Our dataset is licensed under the Apache License 2.0.

\section*{Acknowledgements}

The research work described in this paper was conducted in the JC STEM Lab of Machine Learning and Computer Vision funded by The Hong Kong Jockey Club Charities Trust. This research received partially support from the Global STEM Professorship Scheme from the Hong Kong Special Administrative Region.

\bibliography{reference}

\appendix

\section*{Appendix}

\section{Details in Open-end QA-pairs Generation}
\label{a1}

First, for the construction of procedures, we summarized the annotated steps and their order for each task and condensed existing descriptions into a concise textual form, which serves as a procedural guideline for completing that task. For QA-pairs generation, we first standardized the annotations in the datasets. Because the four source datasets we selected all contain erroneous steps and related information, we unified their JSON formats and assigned each task a 'task\_id', which is in one-to-one correspondence with procedural text. Subsequently, messages containing the step description, the erroneous action, and an explanation of the error and its cause were jointly provided to the LLM to generate QA-pairs. To encourage diversity, we required the model to generate two or three QA-pairs for the same step segment. To address the issue of human annotator consistency, three annotators reviewed the generated QA-pairs, removing those that exhibited hallucinations or deviated entirely from assessing the erroneous content. Then, a single annotator added additional QA-pairs that the model had not covered, such as asking whether a step was omitted or whether the order of steps was incorrect. Assigning these more subjective QA additions to one single annotator helps ensure consistency. The guidelines for 3 human annotators are as follows. The guideline for one annotator to generate new QA-pairs is: 'Review the existing QA-pairs in the dataset and, from diverse perspectives, construct additional QA-pairs centered on error detection.'

It is worth noting that the questions in EgoErrorVQA are highly detailed, targeting specific actions and attributes such as the tools used, the amount of ingredients (e.g., the quantity of chopped vegetables), and the heating duration. This design allows us to focus on fine-grained procedural details, making our evaluation both detailed and comprehensive.

We select Qwen2.5-7B-Instruct because it is an open-source model, which enhances the reproducibility of our dataset and facilitates subsequent researchers in expanding the training data. Our generation process is purely text-based, making the dataset more controllable and avoiding interference from irrelevant visual information, as all necessary content is already contained in the text (we have provided the description). We set `max\_new\_tokens = 300` and `do\_sample = False` to ensure the controllability of generation, while encouraging diversity in the generated data by prompting the model to produce two or three distinct QA-pairs. The prompts used to generate are as follows:

\begin{promptbox}[title=System Prompt]
\small "You are a helpful assistant that generates factual QA pairs strictly based on given instructions."
\end{promptbox}

\begin{promptbox}[title=Example for procedure]
\small "Microwave Egg Sandwich":\\

"To prepare a microwave egg sandwich, first coat a 6-oz ramekin with cooking spray, crack an egg into it, and microwave for 30 seconds. Next, cut an English muffin in half with knife, stir the partially cooked egg, and continue microwaving for an additional 15-30 seconds until the egg is almost set. Add 1 tablespoon of salsa and 1 tablespoon of cheese to the egg, microwave briefly until the cheese melts about 10 seconds, then assemble the sandwich by placing the egg on lettuce in the bottom half of the muffin and topping with the other half.",
\end{promptbox}

\begin{promptbox}[title=Prompt for {is\_error=True}]
\small Procedure: \{procedure\}\\ 
Task: \{activity\_name\}\\
Step: \{original\_description\}\\
Error: Yes (Details: \{error\_descriptions\})\\

Generate exactly 2 different English question-answer pair from a first-person perspective (e.g., 'Did I...?''what did I...?'). The question should ask about the error, and the answer must explain what went wrong based on the annotation. The generated questions should be diverse...\\

Respond in JSON format only:\\
\{"qa\_pairs": [\{"question": "...", "answer": "..."\}]\}
\end{promptbox}

\begin{promptbox}
[title=Prompt for {is\_error = False}]
\small Procedure: \{procedure\}\\
Task: \{activity\_name\}\\
Step: \{original\_description\}\\
Error: No (Correct execution)\\

Generate exactly 2 different English question-answer pair from a first-person perspective (e.g., 'Did I...?''what did I...?'). The question should check if the step was done correctly, and the answer should confirm it was done properly...\\

Respond in JSON format only:\\
{"qa\_pairs": [{"question": "...", "answer": "..."}]}.
\end{promptbox}

\begin{promptbox}[title=Review Guidelines]
\small Please conduct a rigorous evaluation for egocentric procedural error detection QA-pairs.\\

PLEASE consider the following dimensions:\\

1. View the corresponding video segment and verify whether its content is consistent with the given QA-pairs.\\

2. For samples containing errors, examine whether the content of the QA-pair is consistent with both the error labels provided in the original dataset and the erroneous behavior exhibited in the video.\\

3. For correct samples, assess whether the QA-pair aligns with the action labels in the original dataset and the actions depicted in the video.\\

4. Examine whether the QA-pair is centered on the verification of procedural actions and error detection, and whether it avoids querying aspects unrelated to the correctness of action execution.\\

Carefully review and remove redundant or non-informative QA-pairs.\\

\end{promptbox}

\section{Why Choose A2A Protocol}
\label{a2}

By building the benchmark on the A2A protocol, our evaluation tasks can be initiated via message-based communication. When researchers wish to evaluate their agents or video-LLMs, they only need to structure their code to accept messages. On public agent platforms such as AgentBeats, many agents and benchmarks adopt this same code organization, which greatly simplifies the evaluation pipeline and enables models to be evaluated on multiple benchmarks without modifying their existing white-agent–style code. Likewise, our benchmark can evaluate a wide range of agents from public platforms, as long as they are implemented in the white agent format.

\section{Details of Experiment and Human Performance Evaluation}
\label{a3}

We selected these four datasets because they are among the few egocentric video datasets that consist entirely of procedural tasks with explicit error annotations and provide action labels, aligning with our needs for dataset construction and benchmarking. Moreover, we carefully examined the selected video segment and confirmed its validity and usability. Assembly101 provides four egocentric camera views, some of which are incomplete. For each video, we therefore use the e3 view by default and resort to e4 only where e3 is missing, as e3 and e4 together cover all required information.

For human performance in open-end VQA, we randomly sampled 75 QA-pairs. Human annotators completed the same tasks as the model, were provided with identical information, viewed the same video clips before answering the questions, and human responses were scored using the same rubric eventually. For human performance in multiple-choice VQA, we randomly sample 100 samples, assign them the same tasks as the model, and compute the corresponding metrics.

The evaluation results of open-end VQA and multiple-choice VQA are presented more intuitively in Table~\ref{tab7}, Fig.~\ref{fig7} and Fig.~\ref{fig8}.

As input for all white agents, in open-end VQA, we provide a procedure text '\{procedure\}', and the corresponding question. In multiple-choice VQA, we provide a procedure text '\{procedure\}', the action label at which the error occurred '\{original\_description\}', the definition of the error category \{error\_type\_definition\}, as well as the corresponding video segment. The model is required to determine whether an error has occurred and, if so, to identify which error category it belongs to. All conditions presented to the model, including the input prompts, are strictly held constant, with the prompt defined as follows.

In both open-end and multiple-choice VQA, we set `max\_new\_tokens = 256` and `do\_sample = False` to ensure output controllability. Experimental results on multiple-choice VQA show that humans achieve state-of-the-art performance on all metrics except Recall. This does not mean humans underperform models, as Recall here denotes the proportion of detected errors. Because the human study uses a small sample, a few missed errors can markedly lower Recall. By contrast, humans obtain the highest precision and F1 scores, capturing error-type classification accuracy and overall error identification, both clearly surpassing model performance.

Qwen’s high Recall mainly stems from over-classifying samples as erroneous, which reduces precision and accuracy. This reflects Qwen’s heightened sensitivity to procedural errors rather than a defect in the evaluation framework.

In our open-end VQA results, we mentioned that Qwen3-VL-8B-Instruct has shown step-by-step reasoning. Here is the case. Step-by-step: 'No, you did not stir the meatball mixture thoroughly before microwaving it. According to the task procedure, you are instructed to stir the mixture **after** the first 1.5 minutes of microwaving, not before. The initial step is to pour the sweet-and-sour sauce mixture over the meatballs and toppings, cover the plate, and microwave for 1.5 minutes then stir and microwave for another 2 minutes. Stirring occurs during the cooking process, not before microwaving begins.'

\begin{figure*}
    \centering
    \includegraphics[width=1\linewidth]{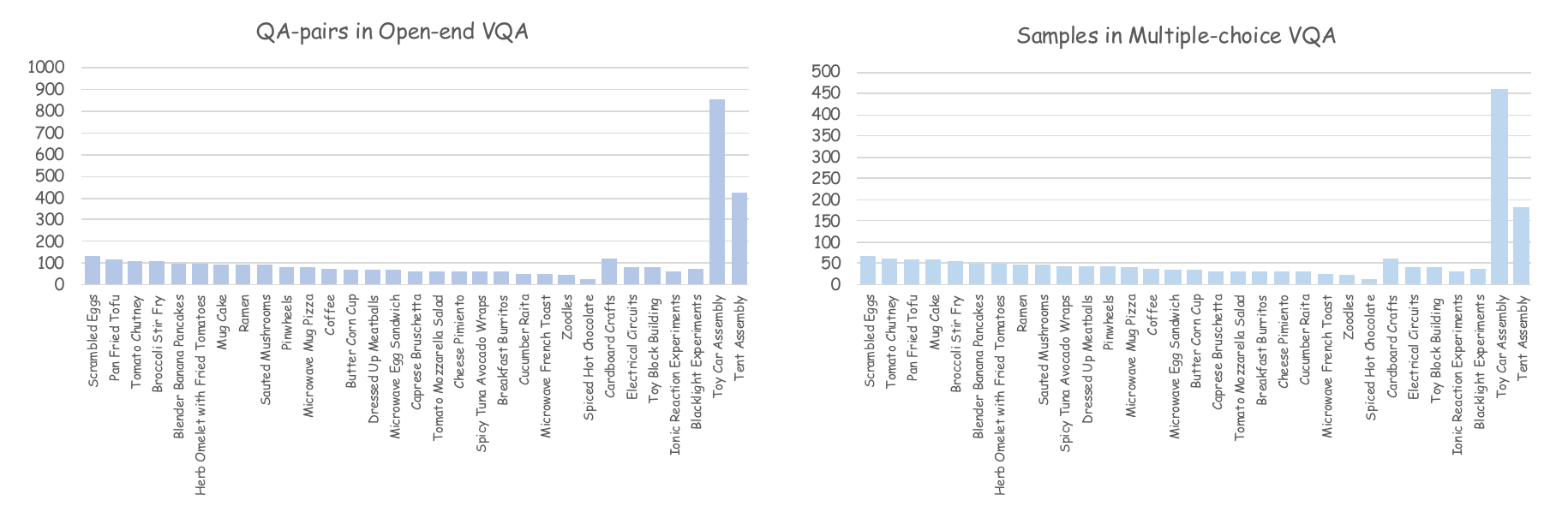}
    \caption{The quantities corresponding to the 31 tasks included in EgoErrorVQA are shown, with the values on the left indicating the number of QA-pairs in Open-end VQA, and the right representing the number of samples, i.e., QA-pairs, in Multiple-choice VQA.}
    \label{fig4}
\end{figure*}
\begin{figure}
    \centering
    \includegraphics[width=0.95\linewidth]{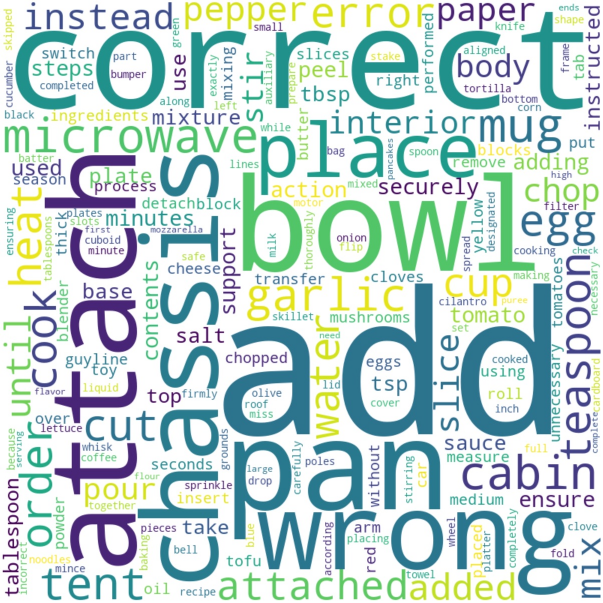}
    \caption{Word cloud of EgoErrorVQA dataset visualizes the relative frequencies of words.}
    \label{fig5}
\end{figure}

\section{Details of Error Type Classification}
\label{a4}

Existing datasets with procedural errors typically define error labels in a scene-specific manner and limit them to the error types observed in that dataset. This heterogeneity in error taxonomies impedes the generalization of downstream methods to new scenarios and complicates the comprehensive evaluation of agents' understanding of procedural errors. For our VQA task, we therefore require a systematic error taxonomy that can cover all potential error types across the selected scenarios. To this end, we exhaustively analyzed all samples from the four chosen datasets and derived a unified error taxonomy that subsumes all identified error types, thereby providing the basis for constructing our multiple-choice VQA evaluation.

Specifically, the detailed definitions of the eight error types are as follows:

• \textbf{Wrong Object}: The operator uses an incorrect tool, material, or component, misuses equipment, or performs incorrect preparation of materials before a step.

• \textbf{Wrong Action}: The operator performs the correct step in an incorrect manner, works in an inappropriate way or position, or makes measurement errors, uses the wrong temperature or cooking time in culinary tasks, or exhibits motor errors when pitching a tent.

• \textbf{Wrong Order}: The operator executes a step before or after its correct position in the sequence. If an action becomes incorrect because a preceding step was already out of order, it is still considered a Wrong Order error due to the propagated ordering mistake.

• \textbf{Omission}: The operator entirely skips a necessary step.

• \textbf{Unintended and Unnecessary Action}: The operator performs an extra step that is not part of the procedure, such as searching for an item while pitching a tent or executing an action that should not occur.

• \textbf{Correct Wrong Action}: The operator recognizes a prior mistake and actively corrects it. This is acceptable behavior but is explicitly annotated as a correction event.

• \textbf{Equipment Failure}: A tool or material fails or malfunctions during the task (e.g., during tent pitching), even when this is not caused by the operator’s error.

• \textbf{Others}: Any error that does not fit the above categories, such as abnormally slow movements.

\begin{figure*}
    \centering
    \includegraphics[width=1\linewidth]{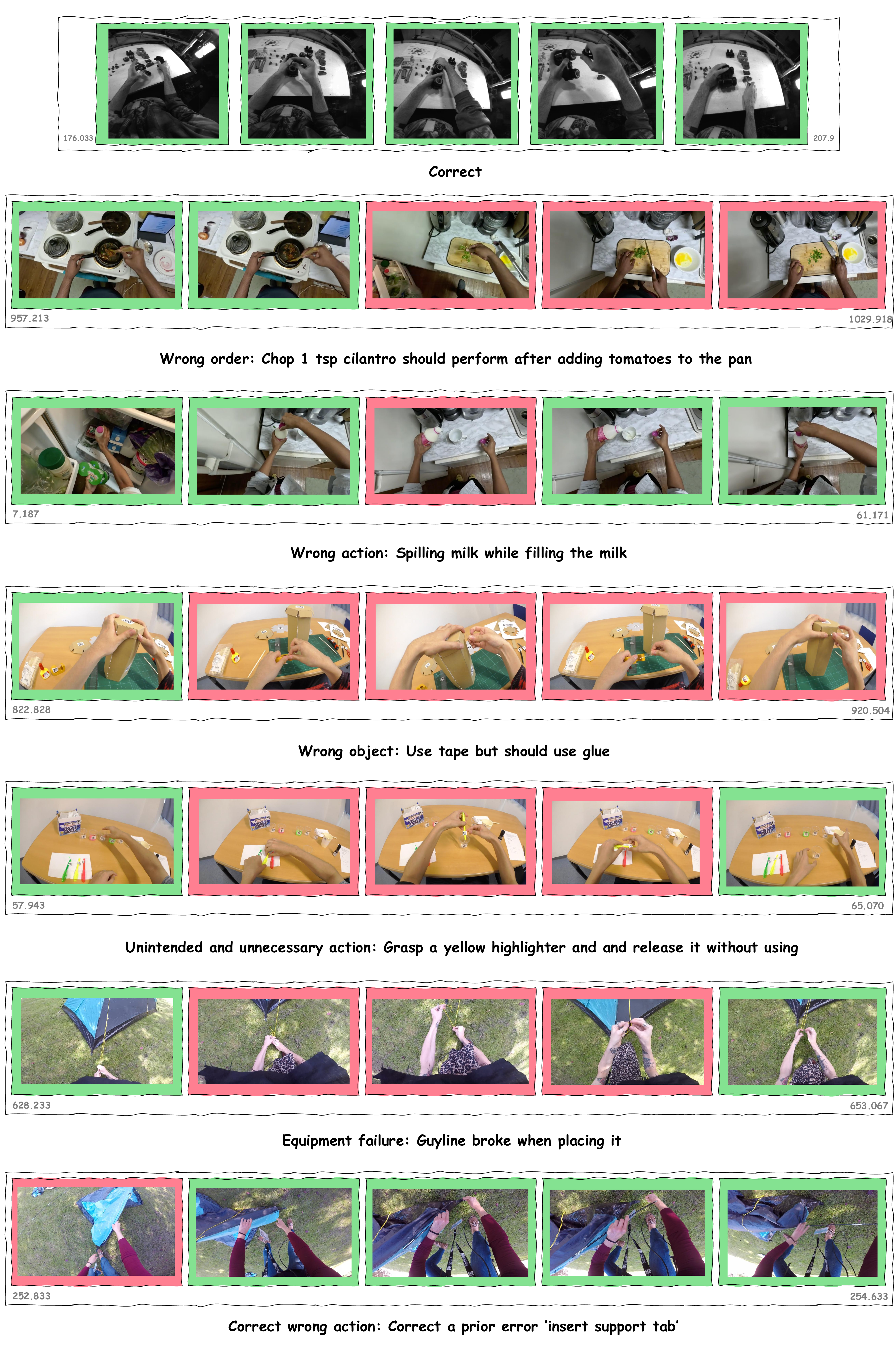}
    \caption{Examples of each Error Type in four selected datasets.}
    \label{fig6}
\end{figure*}

\begin{figure*}
    \centering
    \includegraphics[width=1\linewidth]{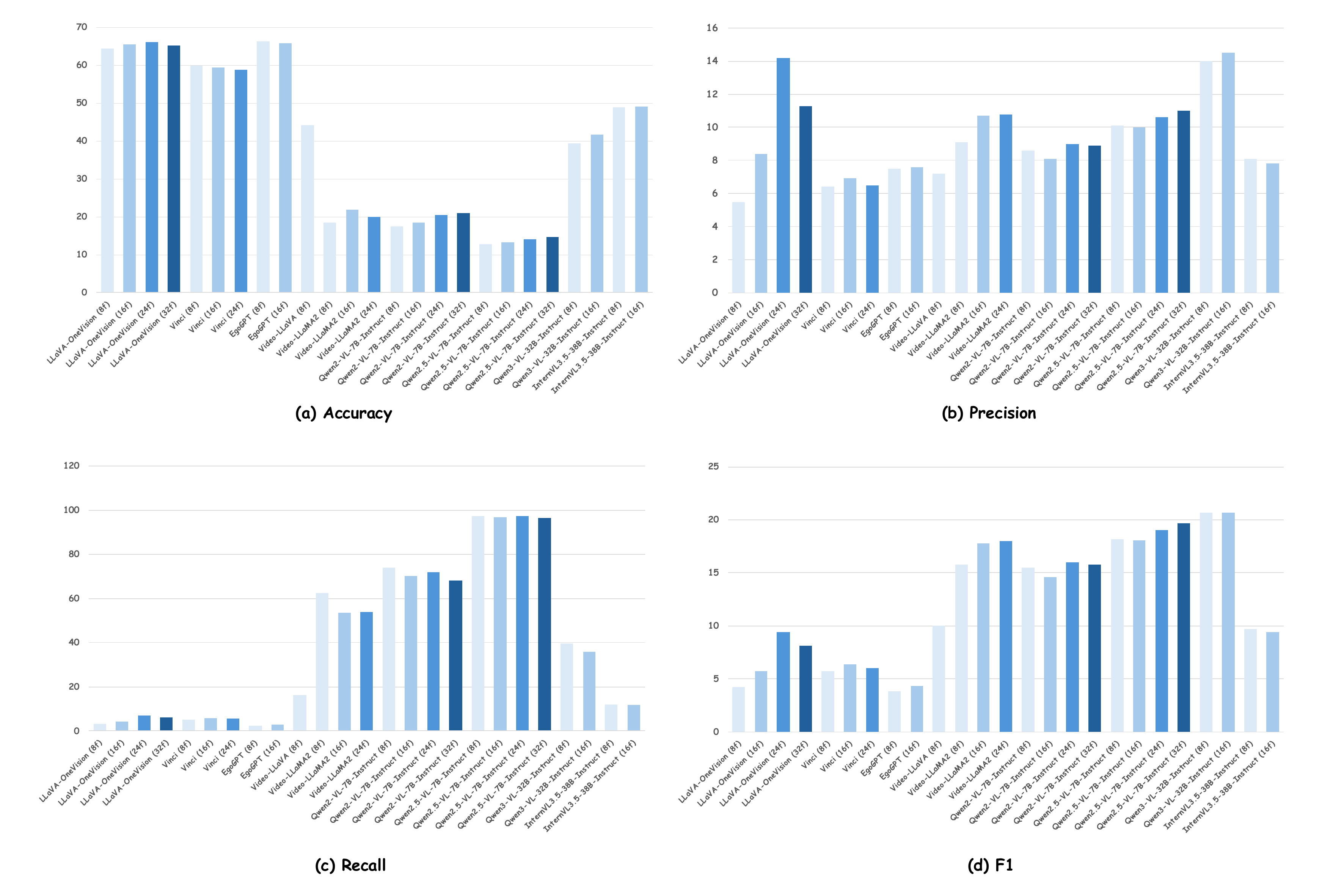}
    \caption{Performance comparison among different agents and models under four metrics in Multiple-choice VQA evaluation.}
    \label{fig7}
\end{figure*}

\begin{figure*}
    \centering
    \includegraphics[width=1\linewidth]{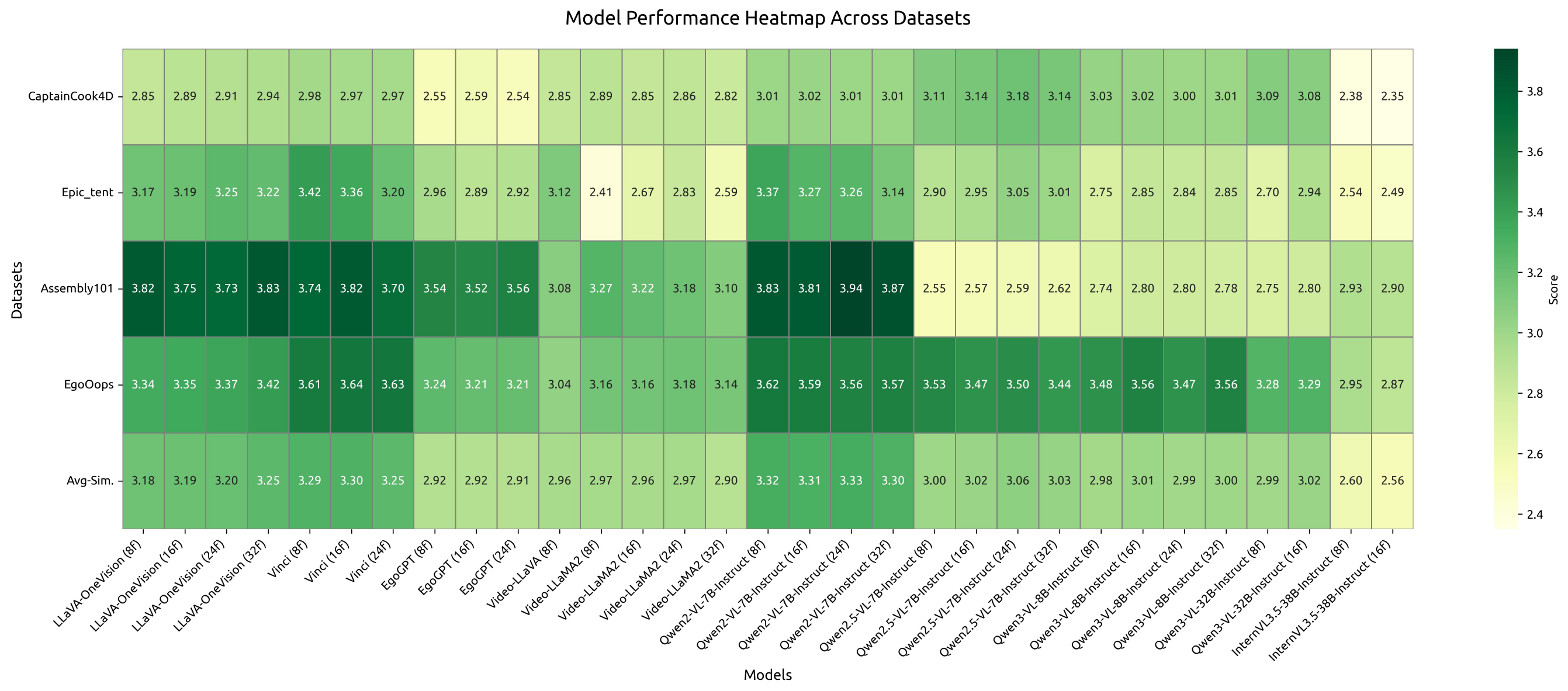}
    \caption{A heatmap of open-end VQA evaluation results.}
    \label{fig8}
\end{figure*}

Examples of each Error Type are shown in Fig.~\ref{fig6}.

Notably, although experiments show generally low precision across models, highlighting the intrinsic difficulty of the task, this is a direct consequence of our benchmark design. Rather than relying on binary detection, correctness requires identifying the precise error type, providing a foundational and pioneering resource for future work on interpretable and fine-grained error analysis.

Furthermore, our error taxonomy is explicitly constructed for fine-grained discriminability. “Wrong order” denotes actions executed in an incorrect sequence while remaining individually correct, whereas “wrong action” captures execution errors within a step, such as using an incorrect heating time or cutting an item into two pieces instead of three. As illustrated by the examples for each category, these definitions yield clearly distinguishable error types.
\begin{promptbox}[title= OE Input]
\small

$<$video\_path$>$\{video\_path\}$<$/video\_path$>$\\
$<$prompt$>$\\

You are an expert procedural error detection assistant. \\
Your task is to carefully analyze egocentric videos and identify any procedural errors by comparing the observed actions with the provided task procedure.\\

Task Procedure: \{procedure\}\\

Question: \{question\}\\

$<$/prompt$>$\\
$<$start\_ts$>$\{start\_time\}$<$/start\_ts$>$\\
$<$end\_ts$>$\{end\_time\}$<$/end\_ts$>$\\
\end{promptbox}

\section{Details of Multiple-choice VQA Metric}
\label{a5}

\textbf{Precision} represents the proportion of samples predicted as procedural error by the model whose error types are correctly matched. \textbf{Accuracy} represents the proportion of correctly predicted samples among all samples of the model. \textbf{Recall} is the proportion of correctly predicted samples among all procedural error samples, and \textbf{F1 Score} is the harmonic mean of precision and recall, it is used to comprehensively evaluate the performance of binary classification models, and is particularly suitable for scenarios with imbalanced datasets. In confusion matrix, we will count $TP$ (True Positive), $FP$ (False Positive), $TN$ (True Negative), and $FN$ (False Negative).
\begin{promptbox}[title= MCQ Input]
\small

$<$video\_path$>$\{video\_path\}$<$/video\_path$>$\\
$<$prompt$>$\\

You are an expert procedural error detection assistant.  \\
Your task is to determine whether the current step contains a procedural error, based on the full task procedure and the expected sequence of actions.\\

Task Procedure: \{procedure\}\\

Current Step: \{original\_description\}\\

Error type definition: \{error\_type\_definition\}\\

Instructions:\\
- Compare the observed action against the full task procedure and step description.\\
- If the action is correct, output ONLY: correct\\
- If there is an error, output ONLY ONE of the error type names above (e.g., Wrong object, Wrong action).\\

$<$/prompt$>$\\
$<$start\_ts$>$\{start\_time\}$<$/start\_ts$>$\\
$<$end\_ts$>$\{end\_time\}$<$/end\_ts$>$\\
\end{promptbox}
The formula definitions of the four evaluation metrics are as follows:
\begin{equation}
\mathbf{Precision}=\frac{TP}{TP+FP},
\end{equation} \begin{equation}
\mathbf{Recall}=\frac{TP}{TP+FN},
\end{equation}
where $TP$ represents the number of samples which model predict they are error, GT are also error, and the error type are match, $FP$ represents the number of samples which model predict are error, but GT are correct or the error type are mismatch.
\begin{equation}
\mathbf{Accuracy}=\frac{TP+TN}{TP+TN+FP+FN},
\end{equation} where $TN$ represents the number of samples which model predict they are correct, GT are also correct. $FN$ represents the number of samples which the model predict they are correct, but GT are some type of error.
\begin{equation}
\mathbf{F1\ Score} = 2 \times \frac{\text{Precision} \times \text{Recall}}{\text{Precision} + \text{Recall}}.
\end{equation}
Also, in simpler terms about confusion matrix, $TN$: Predict is correct, GT is correct, $FP$: Predict is error, but GT is correct or the error type is mismatch,  $FN$: Predict is correct, but GT is some type of error, $TP$: Predict is error, GT is error, and the error type is match.

\section{Details for Caculate Cohen’s Kappa, Pearson and Spearman}
\label{a6}

When computing the Cohen's Kappa and Pearson coefficients, three human annotators scored the same 170 responses using the same criteria applied to Qwen2.5-7B-Instruct and DeepSeek-LLM-7B-Chat. These 170 responses, randomly sampled and representative, covered outputs from both models across four dataset scenarios. As shown in the table, Cohen's Kappa was first computed pairwise between annotators and then aggregated across all three to obtain an overall inter-annotator agreement. This coefficient reflects substantial consensus among annotators, indicating a broadly representative human scoring standard rather than strong individual bias.

\begin{figure*}
    \centering
    \includegraphics[width=1\linewidth]{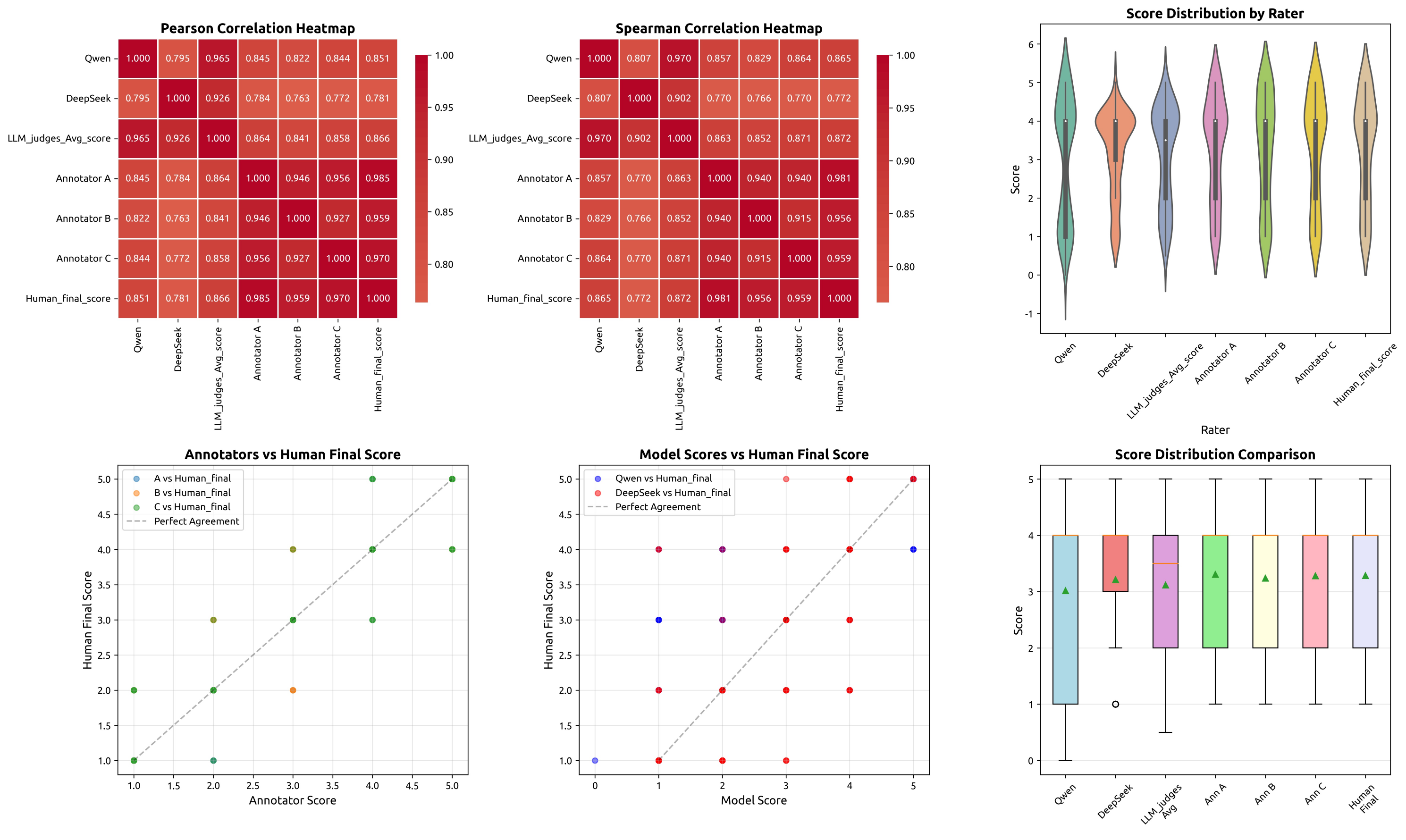}
    \caption{Heatmaps of the Pearson and Spearman correlations, intuitively illustrating the inter-rater relationships. Spearman correlation is a non-parametric statistic that quantifies the strength and direction of a monotonic association between two variables. In addition, it shows the score distribution for each rater, where \textbf{Human Final Score} denotes the final score representing the human evaluation standard obtained via the voting mechanism described above, and \textbf{LLM Judges Avg Score} denotes the average score of the two judge models.}
    \label{fig9}
\end{figure*}

Based on the three annotators’ scores, we then applied a majority-voting scheme: if at least two annotators assigned the same score to a response, that score was taken as the final human rating; if all three scores differed, the annotators discussed the case and reached a consensus score. This procedure yielded a unified human scoring standard. Finally, we computed Pearson correlation coefficients between this unified human rating and the scores produced by Qwen2.5-7B-Instruct and DeepSeek-LLM-7B-Chat, respectively. The resulting coefficients show strong agreement between model scores and the unified human standard, supporting the reliability of the model-based ratings. You can refer to Fig.~\ref{fig9} and Table~\ref{tab8} for detailed visualization results.

In several prior studies, Cohen’s Kappa above 0.8 is regarded as indicating near‐perfect agreement, while values greater than 0.711 are typically interpreted as substantial agreement. Thus, although our inter‐annotator agreement does not reach the near‐perfect level, it still falls within the substantial range and can be taken as representative of general human judgments. Importantly, we do not use Cohen’s Kappa to measure agreement between human evaluators and the LLM; it only reflects the consistency among the three human annotators. We report kappa to demonstrate that the human annotations themselves are reliable and can serve as a robust reference standard. By contrast, Pearson correlation coefficients (0.851 for Qwen and 0.781 for DeepSeek) quantify the alignment between LLM scores and human scores, demonstrating that the LLM judges can be trusted as evaluation proxies.

\begin{table*}
    \centering
    \vspace{-20pt}
    \begin{tabularx}{\textwidth}{XXXX}
    \hline
        {\footnotesize\textbf{Qwen3-VL-8B-Instruct}}  &  {\footnotesize\textbf{Video-LLaVA}}&  {\footnotesize\textbf{EgoGPT}}& {\footnotesize\textbf{Vinci}}\\
         \hline
           {\scriptsize Q: What mistake did I make when using the switch?}\par{\scriptsize A: You connected switch S2 and lamp L1 in parallel and then wired S2 into the series circuit of the motor and battery box...}\par\textcolor{Green}{\scriptsize GT: I used switch S1 instead of switch S2, which is incorrect based on the instructions.}\par\textcolor{purple!45}{\scriptsize Qwen: 2}\par\textcolor{purple!45}{\scriptsize Qwen Reason: The answer describes a different mistake than the one provided in the ground truth. It suggests using S1 instead of S2, which is not mentioned in the model's answer.}\par\textcolor{blue!45}{\scriptsize Deepseek: 4}\par\textcolor{blue!45}{\scriptsize Deepseek Reason: The model's answer is mostly correct, but it could be more clear and concise. It provides a detailed explanation of the mistake, but it could be more concise for a general audience.} & {\scriptsize Q: Did I use the wrong part when I attach track to chassis?}\par{\scriptsize A: Yes, you did. You used a wheel instead of a track.}\par\textcolor{Green}{\scriptsize GT: No, I used the correct part.}\par\textcolor{purple!45}{\scriptsize Qwen: 1}\par\textcolor{purple!45}{\scriptsize Qwen Reason: The model's answer is completely incorrect as it states that a wheel was used instead of a track, while the ground truth indicates that the correct part was used.}\par\textcolor{blue!45}{\scriptsize Deepseek: 2}\par\textcolor{blue!45}{\scriptsize Deepseek Reason: The model's answer is partially correct. It correctly identifies that a wheel was used instead of a track, but it also contains a minor error by stating that the wheel was used instead of a track. The ground truth is that the correct part was used, so the model's answer is mostly wrong.} & {\scriptsize Q: What went wrong when I tried to detach the bumper from the body?}\par{\scriptsize A:The screw was not properly loosened, making it difficult to remove.}\par\textcolor{Green}{\scriptsize GT: This action was unnecessary because the bumper was already properly attached to the body.}\par\textcolor{purple!45}{\scriptsize Qwen: 1}\par\textcolor{purple!45}{\scriptsize Qwen Reason: The model's answer does not match the ground truth. The ground truth states that detaching the bumper was unnecessary, while the model suggests a specific technical issue with the screw.}\par\textcolor{blue!45}{\scriptsize Deepseek: 3}\par\textcolor{blue!45}{\scriptsize Deepseek Reason: The model's answer is partially correct. It correctly identifies the issue with the screw, but it does not fully address the ground truth that the bumper was already properly attached to the body.}  & {\scriptsize Q: Did I discard both ends of the tortilla without any leftovers?}\par{\scriptsize A:You did discard both ends of the tortilla without any leftovers.}\par\textcolor{Green}{\scriptsize GT: Yes, I discarded both ends of the tortilla completely.}\par\textcolor{purple!45}{\scriptsize Qwen: 5}\par\textcolor{purple!45}{\scriptsize Qwen Reason: The model's answer is a perfect match to the ground truth.}\par\textcolor{blue!45}{\scriptsize Deepseek: 4}\par\textcolor{blue!45}{\scriptsize Deepseek Reason: The model's answer is mostly correct, but it lacks the specific detail of discarding both ends of the tortilla completely. It could be interpreted as partially correct or mostly correct.} \\
         \hline
    \end{tabularx}
    \caption{Performances on Open-end VQA is presented along with its scoring results.}
    \label{tab7}
\end{table*}
\begin{table}
    \centering
     \scalebox{0.8}{
    \begin{tabular}{ccccc}
    \hline
         Coefficient&  A vs B&  A vs C&  B vs C& All\\
         \hline
         Cohen's Kappa&  0.716&  0.771&  0.646& 0.711\\
 Fleiss' Kappa& -& -& -&0.710\\
         Pearson&  0.950&  0.959&  0.934& -\\
         \hline
    \end{tabular}
    }
    \caption{Results of three coefficients calculated among annotators A, B, and C. Fleiss’ Kappa represents the overall agreement of the three annotators’ ratings.}
    \label{tab8}
\end{table}

\section{Details for Ego-ADR}
\label{a7}

\begin{promptbox}[title= Deep decoupling]
\small

"Watch this video clip from an instructional/procedural task recorded from a first-person (egocentric) perspective.\\

First, describe what the person is doing in 2-3 sentences. Be specific about visual details: mention colors, shapes, sizes, labels, positions, and quantities of objects. \\

Describe the exact action, the objects/tools involved, how the action is performed, and any notable issues (hesitation, fumbling, searching, undoing, idle, equipment breaking).\\

Do NOT guess the intent or expected outcome. Only describe what you see."
\end{promptbox}

\begin{promptbox}[title= Shallow decoupling]
\small

"Watch the video carefully. The person should be performing this step: "\{expected\_step\}"\\
Expected objects/tools: \{expected\_objects\}\\

Answer these 3 questions about what you ACTUALLY see in the video:\\
1. What object or tool is the person using or holding?\\
2. What action is the person performing? (e.g., cutting, stirring, placing, picking up, etc.)\\
3. Does the action look correct and successful, or did something go wrong? (e.g., dropped, broke, wrong item, redo, idle, searching)\\

Be specific and brief."
\end{promptbox}

In \textbf{Text-only probe}, We adopt a two-stage judgment mechanism. In the first stage, if the model exhibits abnormal behavior on text-only inputs, it is directly assigned a shallow decoupling architecture. The second stage evaluates instruction following and textual reasoning using prompts of the form: “Answer STEP:<number>, and compute the formula <formula>.” Models that correctly follow instructions and complete the computation are then promoted to deep decoupling.

This configuration is not positioned as a core contribution. Our main contributions are: (i) the design of a dedicated benchmark, including dataset construction and task formulation; (ii) comprehensive evaluation experiments; and (iii) a decoupling-based strategy to enhance procedural reasoning. Concretely, the model is first guided to identify the relevant step, then required to provide a description before making a final judgment. Compared with standard Chain-of-Thought (CoT), this form of reasoning enhancement is better aligned with our procedural error detection and classification tasks.

In \textbf{Key-step match}, given the action annotation $a$ from the evaluation query and a procedure bank $\mathcal{P} = \{s_1, s_2, \dots, s_K\}$ containing $K$ structured steps for each task, we identify the most relevant expected step $s^*$ using a two-stage text retrieval approach.

First, we construct a TF-IDF index over all step descriptions in $\mathcal{P}$ using unigram and bigram features. The query annotation $a$ is vectorized with the same vocabulary, and we retrieve the step with the highest cosine similarity:
\begin{equation}
  s^* = \arg\max_{s_i \in \mathcal{P}} \;
  \frac{\mathrm{TF\text{-}IDF}(a) \cdot \mathrm{TF\text{-}IDF}(s_i)}
       {\|\mathrm{TF\text{-}IDF}(a)\| \cdot \|\mathrm{TF\text{-}IDF}(s_i)\|}
\end{equation}
When the cosine similarity falls below a threshold $\tau$ (empirically set to 0.25), we fall back to character-level sequence matching using the difflib, which computes the similarity ratio based on the longest common subsequences. The entire matching process is case-insensitive, and 'scikit-learn' provides a built-in stop-word configuration, enabling the effective extraction and matching of key information.

This deterministic matching step requires no neural inference, operates in constant time per query, and provides: (1)~the matched step description and its associated expected objects/tools, (2)~contextual neighboring steps (previous and next), and (3)~a confidence signal for downstream error classification.

In \textbf{Video Narration}, prompts for \textbf{deep decoupling} and \textbf{shallow decoupling} are defined as follows. Since conventional error detection methods are not applicable to QA scenarios, a direct transfer of such comparisons is not feasible. We therefore adopt the well-established Chain-of-Thought (CoT) strategy for zero-shot performance comparison in our experiments.

The CoT prompt is defined as follows.

\section{Prompt in LLM Judges}
\label{a8}

For both LLM judges, we set `max\_new\_tokens = 256` and `do\_sample = False` to ensure output controllability.

\begin{promptbox}[title=Prompt for Human Judges]
\small You are an expert evaluator. Score the model's answer on accuracy and completeness, evaluate the quality of the model's answer by comparing it with the ground truth.\\
Question: \{question\}\\
Ground Truth: \{ground\_truth\}\\
Model's Answer: \{model\_answer\}\\
Rate the answer on a scale of 0 to 5:\\
{5 = Perfect match, 4 = Minor error, 3 = Partially correct, 2 = Mostly wrong, 1 = Wrong, 0 = No response}\\

Output ONLY: \{"score": int, "reason": "brief explanation"\}
\end{promptbox}

\begin{promptbox}[title=Prompt for Qwen Judge]
\small You are an expert evaluator. Score the model's answer on accuracy and completeness, evaluate the quality of the model's answer by comparing it with the ground truth.\\
Question: \{question\}\\
Ground Truth: \{ground\_truth\}\\
Model's Answer: \{model\_answer\}\\
Rate the answer on a scale of 0 to 5:\\
{5 = Perfect match, 4 = Minor error, 3 = Partially correct, 2 = Mostly wrong, 1 = Wrong, 0 = No response}\\

Output ONLY: \{"score": int, "reason": "brief explanation"\}
\end{promptbox}

\begin{promptbox}[title=Prompt for DeepSeek Judge]
\small You are an expert evaluator. Score the model's answer on accuracy and completeness, evaluate the quality of the model's answer by comparing it with the ground truth.\\
Question: \{question\}\\
Ground Truth: \{ground\_truth\}\\
Model's Answer: \{model\_answer\}\\
Rate the answer on a scale of 0 to 5:\\
{5 = Perfect match, 4 = Minor error, 3 = Partially correct, 2 = Mostly wrong, 1 = Wrong, 0 = No response}\\

Output ONLY: \{"score": int, "reason": "brief explanation"\}
\end{promptbox}

\begin{promptbox}[title=CoT Prompt]
\small ...\\
\#f"Please reason step by step based solely on what is observed in the video and the expected procedure:\textbackslash{}n\textbackslash{}n"\\

\#f"Step 1 - What is actually happening?\textbackslash{}n"\\
\#f" - Describe precisely how the action is performed (e.g., hand motion, tool usage, object interaction).\textbackslash{}n"
\#f" - List the objects involved and the timing relative to other steps.\textbackslash{}n\textbackslash{}n"\#\\
f"Step 2 - What should be happening?\textbackslash{}n"\\
\#f" - According to the full task procedure, what is the correct way to perform this step?\textbackslash{}n"\\
\#f" - What objects should be used, and at what point in the sequence?\textbackslash{}n\textbackslash{}n"\\
\#f"Step 3 - Compare reality vs. expectation:\textbackslash{}n"
\#f" - Are there any discrepancies in how, when, or with what the step is carried out?\textbackslash{}n"\\
\#f" - If everything matches the expected behavior, there is no error.\textbackslash{}n\textbackslash{}n"\\
\#f"Step 4 - Final judgment:\textbackslash{}n"\\
\#f" - If no discrepancy exists, output 'correct'.\textbackslash{}n"\\
\#f" - If a discrepancy exists, determine which single error category (as defined in your guidelines) best captures the core issue.\textbackslash{}n\textbackslash{}n"\\

\#f"Output Format:\textbackslash{}n"\\
\#f"Output ONLY the final error type name. No explanation, no intermediate text.")
\end{promptbox}

\begin{promptbox}[title=Prompt for Building the Procedure Bank]
\small [STRUCTURIZE\_PROMPT = """\\
You are a procedural task analyst. Given a natural language procedure description for a task, decompose it into a structured list of steps.\\

For each step, extract:\\
1. step\_index: integer starting from 1\\
2. action\_description: a concise description of what to do in this step\\
3. tools: list of tools/equipment used (e.g., ["knife", "cutting board"]). Use [] if none.\\
4. objects: list of objects/ingredients involved with quantities if mentioned (e.g., ["onion (1/4 medium)", "garlic (2 cloves)"])\\
5. expected\_state\_change: what changes after this step is done (e.g., "Onion is diced into small pieces")\\
6. original\_text: the exact substring from the original procedure that corresponds to this step\\

Output ONLY valid JSON in this exact format (no markdown, no explanation):\\
\{\{\\
  "task\_id": "<task name>",\\
  "structured\_steps": [\{\{\\
      "step\_index": 1,
      "action\_description": "...",
      "tools": \{"..."\},
      "objects": \{"..."\},
      "expected\_state\_change": "...",
      "original\_text": "..." \}\}]\\
\}\}\\
Task: "\{task\_id\}"\\
Procedure: "\{procedure\_text\}"\\
"""
]
\end{promptbox}

\end{document}